\documentclass[10pt,twocolumn,letterpaper]{article}

\usepackage[pagenumbers]{cvpr} 

\usepackage{graphicx}
\usepackage{amsmath}
\usepackage{amssymb}
\usepackage{booktabs}

\usepackage{mathrsfs}
\usepackage{circledsteps}

\newcommand{\veryshortarrow}[1][3pt]{\mathrel{%
   \hbox{\rule[\dimexpr\fontdimen22\textfont2-.2pt\relax]{#1}{.4pt}}%
   \mkern-4mu\hbox{\usefont{U}{lasy}{m}{n}\symbol{41}}}}

\makeatletter

\setbox0\hbox{$\xdef\scriptratio{\strip@pt\dimexpr
    \numexpr(\sf@size*65536)/\f@size sp}$}

\newcommand{\scriptveryshortarrow}[1][3pt]{{%
    \hbox{\rule[\scriptratio\dimexpr\fontdimen22\textfont2-.2pt\relax]
               {\scriptratio\dimexpr#1\relax}{\scriptratio\dimexpr.4pt\relax}}%
   \mkern-4mu\hbox{\let\f@size\sf@size\usefont{U}{lasy}{m}{n}\symbol{41}}}}

\makeatother

\usepackage[pagebackref,breaklinks,colorlinks]{hyperref}

\usepackage[capitalize]{cleveref}
\crefname{section}{Sec.}{Secs.}
\Crefname{section}{Section}{Sections}
\Crefname{table}{Table}{Tables}
\crefname{table}{Tab.}{Tabs.}

\def\cvprPaperID{8986} 
\def\confName{CVPR}
\def\confYear{2022}

\begin{document}

\title{PolyWorld: Polygonal Building Extraction with Graph Neural Networks in Satellite Images}

\author{Stefano Zorzi $^{1 ~ 3}$\\
{\tt\small stefano.zorzi@icg.tugraz.at}
\and
Shabab Bazrafkan $^2$\\
{\tt\small sbazrafkan@blackshark.ai}
\and
Stefan Habenschuss $^2$\\
{\tt\small shabenschuss@blackshark.ai}
\and
Friedrich Fraundorfer $^1$\\
{\tt\small fraundorfer@icg.tugraz.at}
\and
$^1$ Graz University of Technology\\
$^2$ Blackshark.ai ~ $^3$ VRVis\\
}
\maketitle

\begin{abstract}
While most state-of-the-art instance segmentation methods produce binary segmentation masks, geographic and cartographic applications typically require precise vector polygons of extracted objects instead of rasterized output.
This paper introduces PolyWorld, a neural network that directly extracts building vertices from an image and connects them correctly to create precise polygons.
The model predicts the connection strength between each pair of vertices using a graph neural network and estimates the assignments by solving a differentiable optimal transport problem.
Moreover, the vertex positions are optimized by minimizing a combined segmentation and polygonal angle difference loss.
PolyWorld significantly outperforms the state of the art in building polygonization and achieves not only notable quantitative results, but also produces visually pleasing building polygons.
Code and trained weights are publicly available at \url{https://github.com/zorzi-s/PolyWorldPretrainedNetwork}.
\end{abstract}

\begin{figure}[t]
  \centering
    \includegraphics[width=1\linewidth]{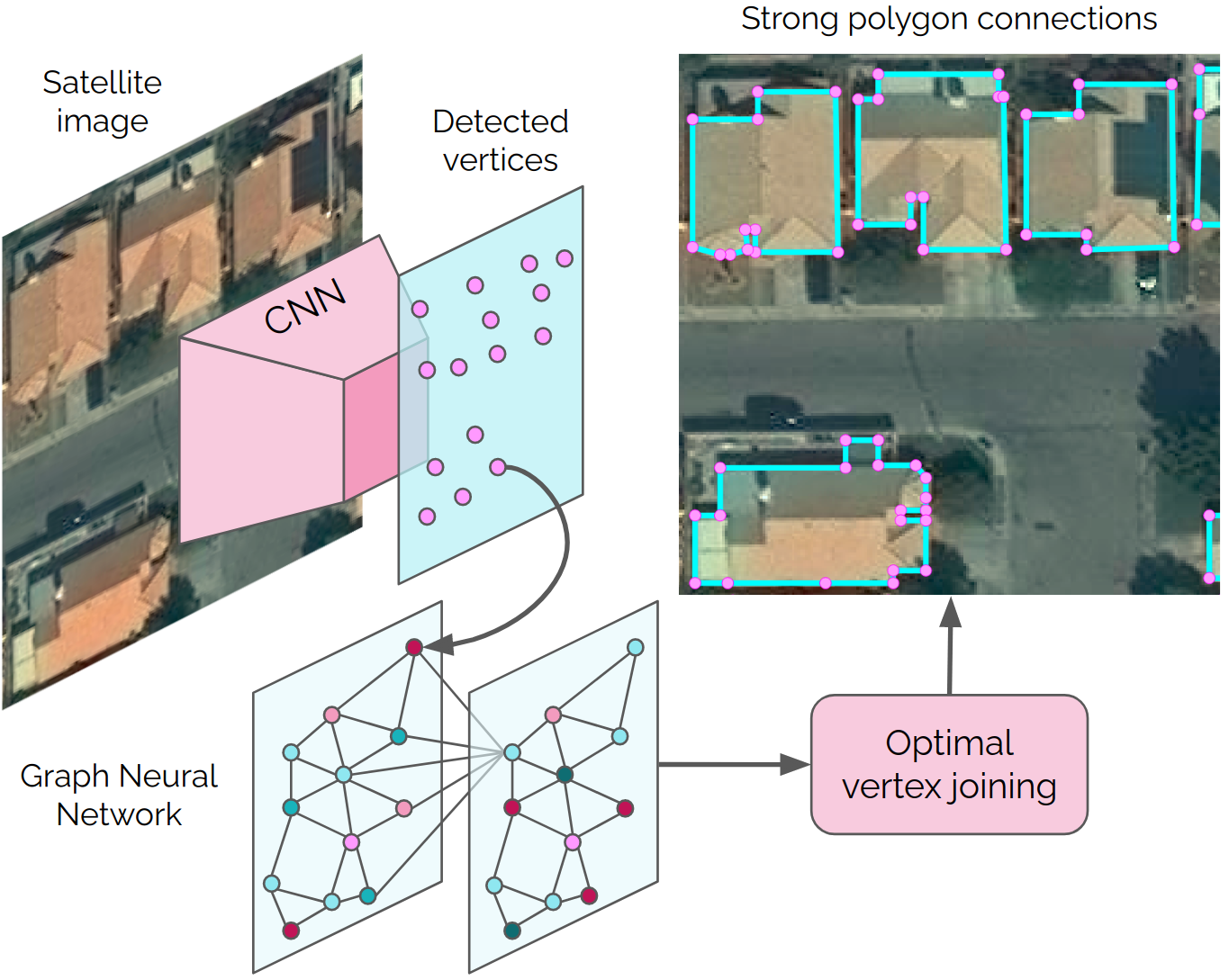}
    \caption{\textbf{Polygonal object extraction with PolyWorld}. The method uses a CNN backbone to detect vertex candidates from an image, and aggregates the information of the visual descriptors exploiting a graph neural network. The connections between vertices are generated solving a differentiable optimal transport problem.}
    \label{fig:first_page}
\end{figure}

\section{Introduction}
The extraction of vector representations of building polygons from aerial and satellite imagery has been growing in importance in many remote sensing applications, such as cartography, city modelling and reconstruction, as well as map generation. 
Most building extraction and polygonization methods rely on the vectorization of probability maps produced by a segmentation network.
These approaches are not end-to-end learned, which means that imperfections and artifacts produced by the segmentation model are carried through the entire pipeline with the consequent generation of unregular polygons.

In this paper, we present a new way of tackling the building polygonization problem.
Rather than learning a segmentation network which is then followed by a polygonization method, we propose a novel neural network architecture called PolyWorld that detects building corners from a satellite image and uses a learned matching procedure to connect them in order to form polygons.
Thereby, our method allows the generation of valid polygons in an end-to-end fashion.

PolyWorld extracts positions and visual descriptors of building corners using a Convolutional Neural Network (CNN) and generates polygons by evaluating whether the connections between vertices are valid.
This procedure finds the best connection assignment between the detected vertex descriptors, which means that every corner must be matched with the subsequent vertex of the polygon.
The connections between polygon vertices can be represented as the solution of a linear sum assignment problem.
In PolyWorld, an important role is played by a Graph Neural Network (GNN) that propagates global information through all the vertex embeddings, increasing the descriptors' distinctiveness.
Moreover, it refines the position of the detected corners in order to minimize the combined segmentation and polygonal angle difference loss.
PolyWorld demonstrates superior performance compared to the state of the art building extraction and polygonization methods, not only achieving higher segmentation and detection results, but also producing more regular and clean building polygons.

\section{Related work}

Since building detection and segmentation from satellite images has been of major research interest throughout the last few decades, discussing all work is beyond the scope of this paper. In this section we therefore focus on the most relevant contributions in different related categories. \smallskip

\textbf{Building segmentation:} Before the great success of deep learning methods, building footprint delineation was mainly done  with multi-step, bottom-up approaches by combining multi-spectral overhead images and airborne LIDAR data\cite{sohn2007data, awrangjeb2010automatic}. 
Nowadays, deep learning-based methods are state-of-the-art, mainly addressing the problem by refining raster footprints via heuristic polygonization approaches computed by powerful semantic or instance segmentation networks \cite{hamaguchi2018building, iglovikov2018ternausnetv2, golovanov2018building, iglovikov2018ternausnet, liu2018path, he2017mask}.
The majority of these segmentation models are trained with cross entropy, soft intersection over union, or Focal based losses \cite{lin2017focal, berman2018lovasz, rahman2016optimizing, sudre2017generalised}, achieving high scores in terms of intersection over union, recall, and precision, but mostly generating irregular building outlines that are neither visually pleasing, nor employable in most cartographic applications.
A typical problem of semantic and instance segmentation networks is, in fact, the inability of outlining straight building walls and sharp corners in presence of ground truth noise, e.g. misalignment between a segmentation mask and an intensity image.
Some publications, therefore, suggest to post-process the segmented building footprints in order to align the segmentation outlines to the actual building contours described in the intensity image.
DSAC \cite{marcos2018learning} employs an Active Contour Model to integrate geometrical priors and constraints in the segmentation process, while DARNet \cite{cheng2019darnet} proposes a loss function that encourages the contours to match the building boundaries.
Another technique to make the building contours more regular and realistic is to combine adversarial and regularized losses \cite{zorzi2019regularization, tang2018normalized, tang2018regularized}. \smallskip

\textbf{Polygon prediction:} Standard semantic and instance segmentation networks are easy to train and generate accurate segmentation masks, but most remote sensing applications that involve building layers require segmentation data in vector format rather than rasterized masks.
Object detection and polygonization methods found in literature can be classified into two categories.

The first category includes methods that perform the vectorization of grid-like information, e.g. the probability map produced by a segmentation network.
In \cite{zhao2018building} the authors corrected the segmentation masks produced with Mask R-CNN\cite{he2017mask} by first simplifying the detected boundaries using the Douglas-Peucker algorithm \cite{douglas1973algorithms} and subsequently refining the resulting polygons using a Minimum Descriptor Length method \cite{sohn2012implicit}.
More recently, Chen et al. \cite{chen2021quantization} suggested to regularize the segmentation produced with a CNN via quantizing the histogram of building boundaries in angle space, which can be achieved by exploiting a Relative Angle Gradient Transform.
Zorzi et al. \cite{zorzi2021machine} applied three different networks in series to perform the extraction and polygonization. 
Their method uses a CNN to generate building segmentation; then, it performs a regularization on the raster data by applying an autoencoder trained with regularized \cite{tang2018normalized, tang2018regularized} and adversarial losses, and finally detects building corners using a third CNN.
The polygonization is performed by ordering the detected corners following the regularized boundaries.
All these methods are developed with the idea of decomposing the building extraction and polygonization problem into smaller tasks that can be tackled individually.
As a result, most of these approaches are computationally heavy, they lack of parallelization and their hyperparameters must be carefully tuned in order to achieve the desired results.
Most importantly, since they are composed of a sequence of blocks, these methods can accumulate errors through their pipeline, which can harm the quality of the final polygonization.
The current state of the art in the field is achieved by the Frame Field Learning (FFL) method \cite{girard2021polygonal}, which generates a vector field that encodes useful boundary information alongside the corresponding segmentation mask.
Moreover, the contour is optimized to be aligned to the frame field using an Active Skeleton Model.

The second category is represented by methods that directly learn a vector representation.
PolyTransform \cite{liang2020polytransform} initializes a polygon for every object instance and refines the vertex positions using a Transformer network \cite{vaswani2017attention}.
Curve GCN \cite{ling2019fast} learns a graph convolutional network to deform polygons in an iterative manner.
Some networks also utilize recurrent neural networks (RNN) to extract polygons vertex by vertex, e.g. Polygon-RNN\cite{castrejon2017annotating} and Polygon-RNN++ \cite{acuna2018efficient}.
Also PolyMapper \cite{li2019topological} applies a RNN to predict building and road vertices one by one.
All these methods directly process polygon parameters but they are typically more difficult to train and they need multiple iterations during inference.
Moreover they have troubles dealing with complex building shapes, e.g. structures having curved walls or holes in their shape.
PolyWorld, which is presented in this paper, fits well into the second category of direct polygon prediction, although the employed architecture and general idea fundamentally differs from all existing work.

\section{The PolyWorld Architecture}
\label{sec:architecture}

\begin{figure}[t]
  \centering
    \includegraphics[width=1\linewidth]{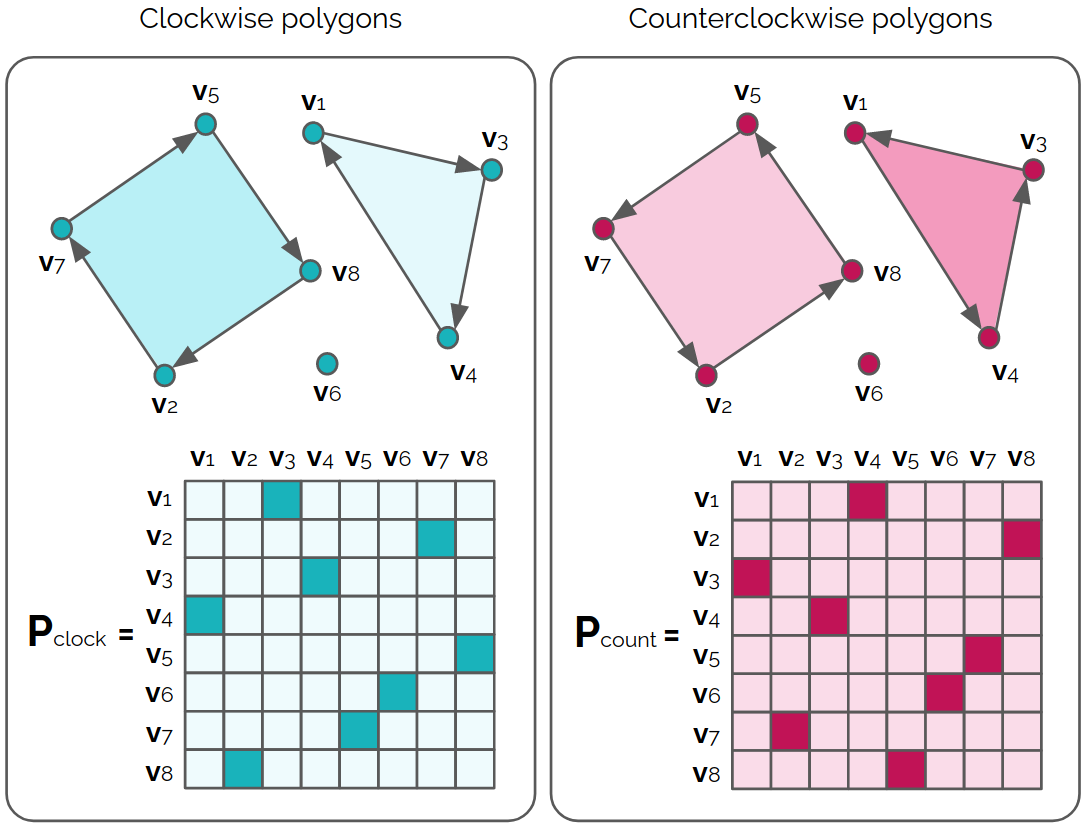}
    \caption{In PolyWorld, the connections between polygon vertices are described with a permutation matrix. The $i$-th row of the permutation matrix $P_{clock}$ or $P_{count}$ indicates the index of the next clockwise or counterclockwise vertex connected to $v_i$. Please note that the permutation matrix of the clockwise oriented polygons $P_{clock}$ is the transpose of the permutation matrix of the counterclockwise oriented polygons $P_{count}$.}
    \label{fig:polygons}
\end{figure}

The main idea behind PolyWorld is to represent building polygons in the scene as a set of vertices connected according to a permutation matrix, as illustrated in Figure \ref{fig:polygons}.
Each corner of the polygon is associated to a specific row of the permutation matrix that indicates the next clockwise vertex.
The permutation matrix must fulfill certain polygonal constraints: \textcolor{black}{\Circled{1}} every vertex corresponds to at most one clockwise connection and one counterclockwise connection; \textcolor{black}{\Circled{2}} the permutation matrix of the clockwise oriented polygons is the transpose of the counterclockwise permutation matrix; \textcolor{black}{\Circled{3}} a vertex having its entry in the diagonal of the permutation matrix can be discarded since, in reality, there are no building polygons having a single corner, e.g. vertex $v_6$ in Fig. \ref{fig:polygons}.

PolyWorld is composed of three blocks: a Vertex Detection Network that extracts a set of possible building corner candidates, an Attentional Graph Neural Network that aggregates information through the vertices and refines their position, and an Optimal Connection Network that generates the connections between vertices.
Given the input image, the model provides the position of the detected building corners and a valid permutation matrix.

\subsection{Vertex Detection Network}

\begin{figure}[t]
  \centering
    \includegraphics[width=1\linewidth]{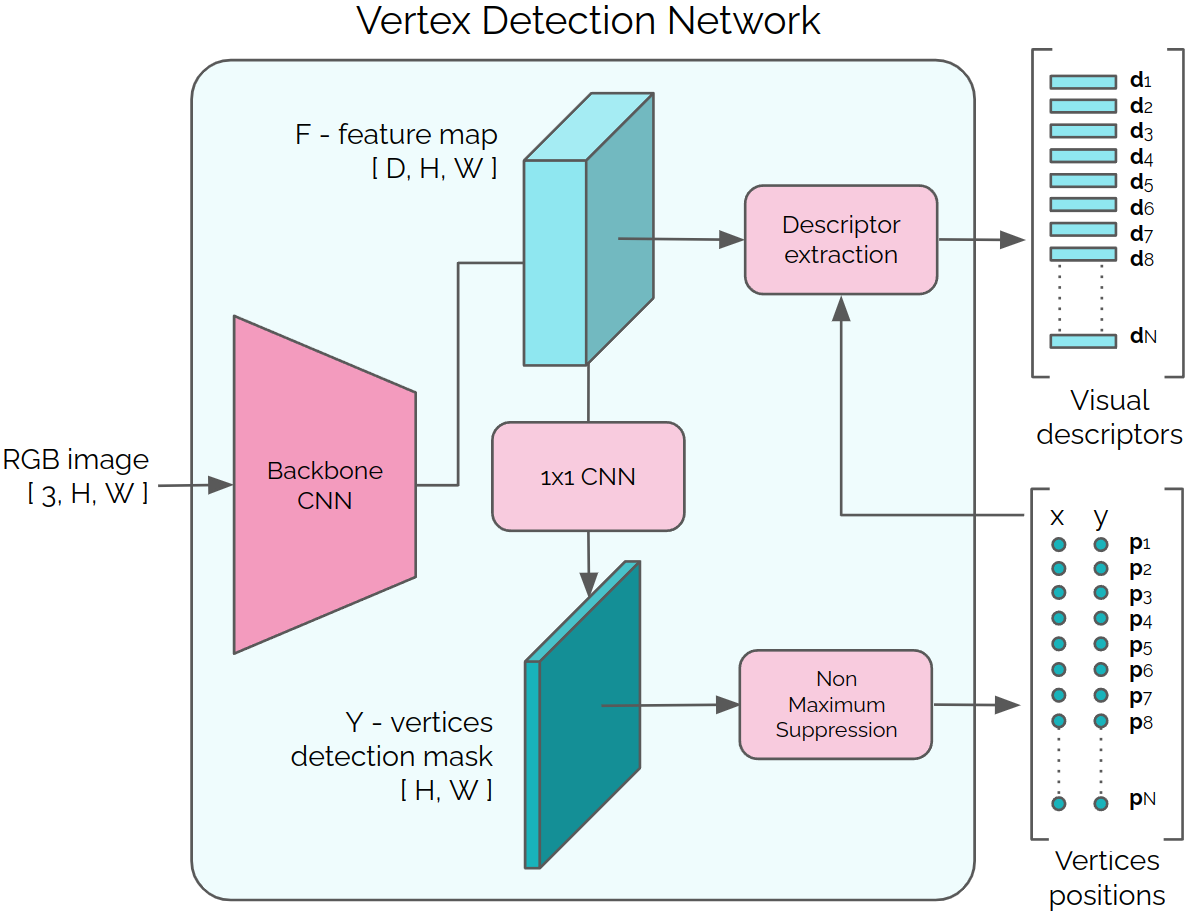}
    \caption{The Vertex Detection Network of PolyWorld. A backbone CNN receives the intensity image and returns a feature map and a vertex detection mask. A Non Maximum Suppression (NMS) algorithm removes undesired vertices and returns $N$ locations that correspond to the highest peaks in the detection mask. The visual descriptors are then extracted from the feature map at every location provided by the NMS.}
    \label{fig:vertices_detection}
\end{figure}

The vertex detection network is depicted in Figure \ref{fig:vertices_detection}.
The module receives an image $I \in \mathbb{R}^{3 \times H \times W}$ as input, it forward propagates $I$ through a fully convolutional backbone, and returns a $D$-dimensional feature map $F \in \mathbb{R}^{D \times H \times W}$.
The vertex detection mask $Y \in \mathbb{R}^{H \times W}$ is obtained by propagating the features $F$ through a $1 \times 1$ convolutional layer.
The detection mask $Y$ is then filtered using a Non Maximum Suppression algorithm with kernel size of 3, in order to retain the most relevant peaks.
The \textit{positions} $p$ of the $N$ highest peaks are then used to extract $N$ \textit{visual descriptors} $d \in \mathbb{R}^D$ from the feature map $F$.
Vertex positions consist of $x$ and $y$ image coordinates $p_i:=(x,y)_i$.
During training, the backbone not only learns to produce a feature map $F$ useful to segment building corners but also learns to embed an abstract representation of the latter.
During training, this information is constrained to represent the building vertex by matching with the other detected corners.

\subsection{Attentional Graph Neural Network}

\begin{figure*}
  \centering
    \includegraphics[width=1\linewidth]{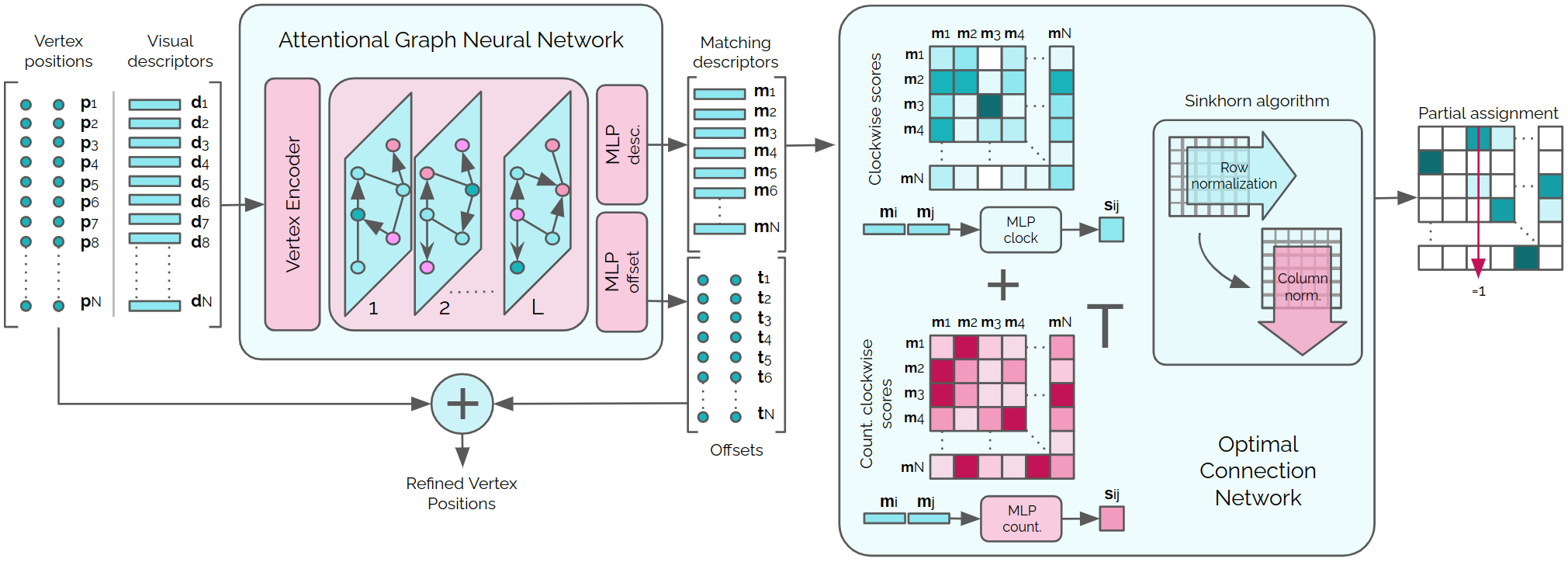}
    \label{fig:short-a}
  \caption{The Attentional Graph Neural Network and the Optimal Connection Network of PolyWorld. The first module uses a vertex encoder to map \textit{vertex positions} $p$ and \textit{visual descriptors} $d$ into a single vector, and uses $L$ self-attention layers to increase their distinctiveness. The module returns a set of \textit{offsets} $t$ and the \textit{matching descriptors} $m$. The offsets are used to refine the vertex positions, while $m$ are propagated through the optimal connection network that creates a $N \times N$ score matrix and generates the permutation matrix using the Sinkhorn algorithm.}
  \label{fig:short}
\end{figure*}

Besides the position and the visual appearance of a building corner, considering other contextual information is essential to describe it in a more rich and distinctive way.
Capturing relationships between its position and appearance with other vertices in the image can be helpful to link it with corners having the same roof style, having a compatible shape and pose for the matching, or simply with adjacent corners.
Motivated by this consideration, we design the next PolyWorld block using an attentional graph neural network that computes a set of \textit{matching descriptors} $m_i \in \mathbb{R}^D$ by learning short and long term vertex relationships given the vertex positions $p$ and the visual descriptors $v$ extracted by the vertex detection network.
Moreover, this block also estimates a \textit{positional offset} $t_i \in \mathbb{R}^2$ in order to refine the vertex positions, optimizing the corner angle and the footprint segmentation.
As we will show in the following chapters, aggregating features from all the detected vertices and refining the vertex positions leads not only to improved segmentation scores, but also to more realistic building polygons.

\subsubsection{Vertex Encoder}
Before forward propagating through the graph neural network, positions $p$ and visual descriptors $d$ are merged by a Multilayer Perceptron (MLP).

\begin{equation}
    d'_i = \textsc{MLP}_{enc} \left( \left[ d_i || p_i \right] \right)
\end{equation}

$\textsc{MLP}_{enc}$ receives the concatenation $[\cdot||\cdot]$ of $p_i$ and $d_i$ and returns a new descriptor $d'_i \in \mathbb{R}^D$ that encodes positional and visual information together.

\subsubsection{Self Attention Network}
The aggregation is performed by a self-attention mechanism \cite{vaswani2017attention} that propagates the information across vertices, increasing their contextual information.

Given the intermediate descriptors $x \in \mathbb{R}^{D \times N}$, the model employs a linear projection to produce a query $Q(x)$, a key $K(x)$, and a value $V(x)$. 
The weights between the nodes are computed taking the softmax over the dot product $Q(x) K(x)^\top$.
The result is then multiplied with the values $V(x)$ in order to propagate the information across all the vertices.
The attention mechanism can be written as:

\begin{equation}
    A = \text{softmax} \left( \frac{Q(x) \cdot K(x)^\top}{\sqrt{n_k}} \right) V(x)
\end{equation}

\noindent
where the normalization term $n_k$ is the dimension of the queries and keys.

This operation is repeated for a fixed numbers of layers $L$. 
The message $A^{(l)} \in \mathbb{R}^{D \times N}$ is the attention result at layer $l$ and it is used to update the vertex descriptors at every step.
We indicate $a^{(l)}_i$ the $i$-th column of $A^{(l)}$, that represents the attention message relative to the $i$-th vertex of the graph.
In every layer the vertex descriptors are updated as follows:

\begin{equation}
    x^{(l+1)}_i = \textsc{MLP}^{(l)} \left( \left[ x^{(l)}_i || a^{(l)}_i \right] \right)
\end{equation}

The embeddings received by the the first attention layer are the descriptors produced by the vertex encoder $d' = x^{(l=1)}$.
Finally, the embedding of the $i$-th vertex produced by the last attention layer $x^{(L)}_i$ is decomposed in two components: a \textit{matching descriptor} $m_i \in \mathbb{R}^{D}$ and a \textit{positional offset} $t_i \in \mathbb{R}^{2}$.

\begin{equation}
    m_i = \textsc{MLP}_{match}  \left( x^{(L)}_i \right)
\end{equation}

\begin{equation}
    t_i = \textsc{MLP}_{offset}  \left( x^{(L)}_i \right)
\end{equation}

The matching descriptors are used further to generate a valid combination of connections between the vertices, while the offsets are combined with the vertex positions as follows:

\begin{equation}
    \hat{p}_i = p_i + \gamma \cdot t_i
    \label{eq:offset}
\end{equation}

\noindent
where $\gamma$ is a factor that regulates the correction radius since the offsets are generated through a HardTanh activation function and the values range between $-1$ and $1$.

\subsection{Optimal Connection Network}

The last block of PolyWorld is the optimal connection network that connects the vertices generating a permutation matrix $P \in \mathbb{R}^{N \times N}$.
The assignment can be obtained by calculating a score matrix $S \in \mathbb{R}^{N \times N}$ for all possible vertex pairs and maximizing the overall score $\sum_{i,j} P_{i,j} S_{i,j}$.

Given two matching descriptors $m_i$ and $m_j$ encoding the information of two distinct vertices, we exploit $\textsc{MLP}_{clock}$ to detect whether the clockwise connection $m_i \xrightarrow{} m_j$ is possible.
The network receives the concatenation of the two descriptors and returns a high score value if the connection between them is strong; e.g. if $m_i$ represents the top-left corner of an orange roof, it is likely that $m_j$ is the next clockwise vertex if it represents a top-right corner of an orange roof.

\begin{equation}
    s^{clock}_{i \xrightarrow{} j} = \textsc{MLP}_{clock} \left( \left[ m_i || m_j \right] \right)
\end{equation}

Vice versa we estimate how strong is the counterclockwise connection $m_i \xrightarrow{} m_j$ exploiting a second network $\textsc{MLP}_{count}$.

\begin{equation}
    s^{count}_{i \xrightarrow{} j} = \textsc{MLP}_{count} \left( \left[ m_i || m_j \right] \right)
\end{equation}

By enforcing constraint \textcolor{black}{\Circled{2}}, we can establish a consistency check between the clockwise and the counterclockwise path of vertices.
The final score matrix $S$ is calculated as the combination of the clockwise score matrix $S_{clock}$ and the transpose version of the counterclockwise score matrix $S_{count}$:

\begin{equation}
    S = S_{clock} + S^{\top}_{count}
\end{equation}

The double path consistency ensures to have stronger matches, better connections and, ultimately, higher polygon quality.

As a final step, we use the Sinkhorn algorithm \cite{sinkhorn1967concerning, cuturi2013sinkhorn, peyre2019computational, sarlin2020superglue} to find the optimal assignment matrix $P$ given the score matrix $S$.
The Sinkhorn is a GPU efficient and differentiable version of the Hungarian algorithm \cite{munkres1957algorithms}, used to solve linear sum assignment problems, and it consists of normalizing rows and columns of $\text{exp}(S)$ for a certain amount of iterations.

\section{Losses}

\textbf{Detection:} We train the corner detection as a segmentation task using weighted binary cross-entropy loss:

\begin{equation}
\begin{aligned}
\mathcal{L}_{det} =  &- \omega \cdot \sum^H_{i=1} \sum^W_{j=1} \bar{Y}_{i,j} \cdot \text{log} \left(  Y_{i,j}  \right) \\
&- \sum^H_{i=1} \sum^W_{j=1} (1 - \bar{Y}_{i,j}) \cdot \text{log} \left(  1 - Y_{i,j}  \right)
\end{aligned}
\label{eq:detection_loss}
\end{equation}

The ground truth $\bar{Y}$ is a sparse array of zeros. 
Pixels indicating the presence of a building corner have a value of one.
Since the segmentation is heavily unbalanced for the foreground pixels, we use the factor $\omega$ to counterbalance positive samples.\smallskip

\textbf{Matching:} The attentional graph neural network and the optimal connection network of PolyWorld are fully differentiable which allows us to backpropagate from the generated partial assignment to the backbone that generates the visual descriptors.
This path is trained in a supervised manner from the ground truth permutation matrix $\bar{P}$ using cross entropy loss:

\begin{equation}
\begin{aligned}
\mathcal{L}_{match} = - \sum^N_{i=1} \sum^N_{j=1} \bar{P}_{i,j} \cdot \text{log} \left(  P_{i,j}  \right)
\end{aligned}
\end{equation}

Due to the iterative normalization through rows and columns made by the Sinkhorn algorithm, minimizing the negative log-likelihood of the positive matches of $P$ leads to simultaneously maximizing the precision and the recall of the matching.\smallskip

\textbf{Positional refinement:} Due to low image resolution, ground truth misalignments, or wrong building labelling, the position of the vertices provided by the vertex detection network is not optimal in practice.
The subsequent matching procedure, therefore, could produce polygons having corner angles different from the ground truth, altering the visual appeal of the extracted polygons.
In order to repress this phenomenon, we minimize the difference between the corner angles of the predicted polygons and the ground truth polygons.


We indicate with $\mathcal{C}$ the function that converts a permutation matrix and vertex positions to a list of polygons $\mathcal{P}$.
The polygons predicted by PolyWorld and the ground truth polygons are then $\mathcal{P} = \mathcal{C} \left(  \hat{p}, P  \right)$ and $\mathcal{\bar{P}} = \mathcal{C} \left(  \bar{p}, \bar{P}  \right)$, respectively.
Indicating with $\mathcal{P}_k$ the $k$-th polygon instance extracted from the image and composed of a set of clockwise ordered vertex positions, we formulate the angle loss as:

\begin{equation}
\begin{aligned}
    \mathcal{L}_{angle} = \sum^K_{k=1} \sum_{(u\scriptveryshortarrow v\scriptveryshortarrow w)} 1 - \text{exp} \left(  -\sigma \cdot  \left|   \Delta_{k, (u,v,w)}    \right|  \right)\\
    \Delta_{k,(u,v,w)} = \angle  \left(   \hat{p}_u, \hat{p}_v, \hat{p}_w   \right)_k  -  \angle  \left(   \bar{p}_u, \bar{p}_v, \bar{p}_w \right)_k
\end{aligned}
\label{eq:angle_loss}
\end{equation}

\noindent
where $(u \veryshortarrow v \veryshortarrow w)$ denotes the indices of any three consecutive vertices in polygon $\mathcal{P}_k$ and $\mathcal{\bar{P}}_k$. 
The strength of the loss term is regulated by the factor $\sigma$, while $ \angle  \left(   \hat{p}_u, \hat{p}_v, \hat{p}_w   \right)_k$ and $\angle  \left(   \bar{p}_u, \bar{p}_v, \bar{p}_w \right)_k$ indicate the angle at the $v$-th vertex of the polygon $\mathcal{P}_k$ and $\mathcal{\bar{P}}_k$, respectively.

Even if the network is encouraged to fix corner angles, $\mathcal{L}_{angle}$ potentially induces unexpected modifications of the polygon shapes since it leaves some degrees of freedom to the network on how to warp the vertices.
In our experiments the network stretched the polygons in undesired ways while respecting the angle criterion, potentially producing misaligned footprints.
PolyWorld fixes this issue by minimizing a segmentation loss between the ground truth and predicted polygons.
This refinement loss not only inhibits unwanted effects of $\mathcal{L}_{angle}$, but it also increases segmentation scores as documented in the next sections.

We generate the footprint mask of the predicted polygons exploiting a Differentiable Polygon Rendering method \cite{stekovic2021montefloor}.
It is the soft version of the winding number algorithm, that checks whether a pixel location $x$ is inside the polygon $\mathcal{P}_k$ with the equation:

\begin{equation}
\begin{aligned}
    W(x, \mathcal{P}_k) = \sum_{(u \scriptveryshortarrow v)} \frac{\lambda \cdot \text{det}(\overline{\hat{p}_u x}, \overline{\hat{p}_v x} )_k }{1 + \left|  \lambda \cdot \text{det}(\overline{\hat{p}_u x}, \overline{\hat{p}_v x} )_k  \right|} \cdot \angle  \left(   \hat{p}_u, x, \hat{p}_v  \right)_k
\end{aligned}
\label{eq:rasterization}
\end{equation}

\noindent
where $(u \veryshortarrow v)$ are the indices of any two consecutive vertices of $\mathcal{P}_k$, $\text{det}(\cdot)$ is the determinant of vectors $\overline{\hat{p}_u q}$ and $\overline{\hat{p}_v q}$, and the value $\lambda$ fixes the smoothness of the raster contours.

Calculating the winding number for every pixel location in the image, we generate the raster mask $M_k \in \mathbb{R}^{H \times W}$ of the polygon $\mathcal{P}_k$.
The segmentation loss $\mathcal{L}_{seg}$ is finally calculated as the soft intersection over union \cite{rahman2016optimizing} between the ground truth segmentation mask $\bar{M}$ and the combination of extracted polygon masks:

\begin{equation}
\begin{aligned}
    \mathcal{L}_{seg} = \text{softIoU} \left(   \sum^K_{k=1} M_k, \bar{M}    \right)
\end{aligned}
\end{equation}

Since the NMS block is not differentiable, the only way for the network to minimize $\mathcal{L}_{seg}$ and $\mathcal{L}_{angle}$ is to generate a proper set of offsets $t$ for Equation \ref{eq:offset}.

\begin{figure*}[t]
  \centering
    \includegraphics[width=0.99\linewidth]{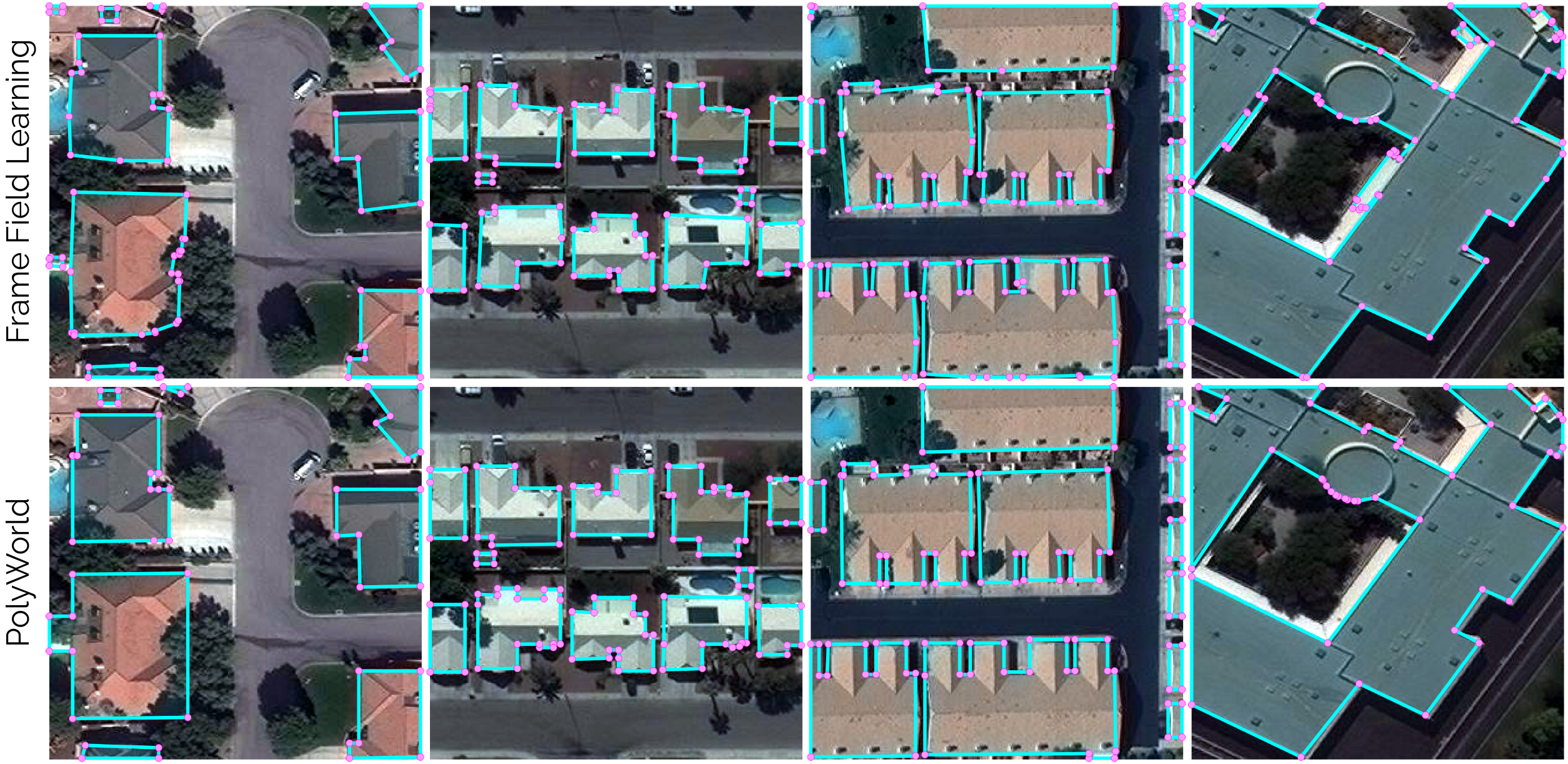}
  \caption{Examples of building extraction and polygonization on CrowdAI test dataset. Top row: Frame Field Learning \cite{girard2021polygonal}} approach with Res101-UNet as backbone and ACM polygonization. Bottom row: PolyWorld results.
  \label{fig:thumbnails}
\end{figure*}

\section{Implementation details}

\textbf{Training and inference:} The NMS algorithm extracts a list of $N=256$ vertex positions $p$ with the highest detection confidence.
During training, these positions are not directly used to extract the descriptors $d$ from the features $F$, but they are first sorted to match the nearest neighboring ground truth point.
After sorting, $p_i$ is the closest vertex to the ground truth point $\bar{p}_i$.
This procedure ensures to have index consistency between the positions $p$ and the ground truth permutation matrix $\bar{P}$.
In reality, the number of extracted points $N$ is always greater than the number of building corners in the image, therefore the vertices that do not minimize the distance with any of the ground truth points have their entry assigned to the diagonal of $\bar{P}$.
PolyWorld is trained from scratch linearly combining detection, matching and refinement losses: $\mathcal{L}_{det} + \mathcal{L}_{match} + \mathcal{L}_{angle} + \mathcal{L}_{seg}$.
Rather than learning the matching branch at the early training stage, we prefer to first pretrain the vertex detection network only using $\mathcal{L}_{det}$.
When it extracts sufficiently accurate building corners, we keep training the full PolyWorld architecture with the complete loss. 
During inference, vertices that have their entry in the diagonal of the permutation matrix are discarded (constraint \textcolor{black}{\Circled{3}}). \smallskip

\textbf{Architecture:} As backbone PolyWorld uses a Residual U-Net model \cite{alom2019recurrent}. 
The descriptor dimension and the intermediate representations of the attentional graph neural network have the same size $D=64$.
We use $L=4$ self attention layers having $4$ parallel heads each.
In Equation \ref{eq:offset}, we use $\gamma=0.05$, allowing a maximum offset radius of $5\%$ of the image size. 
Increasing $\gamma$ further does not improve the results.
We use $\omega = 100$ and $\sigma=10$ in Equations \ref{eq:detection_loss} and \ref{eq:angle_loss} respectively, while, in Equation \ref{eq:rasterization}, the value of $\lambda$ is set to $10^3$ as suggested in \cite{stekovic2021montefloor}.
During training, the permutation matrix $P$ is calculated by performing $T=100$ Sinkhorn iterations, whereas during inference the exact linear sum assignment result is determined using the Hungarian algorithm on the CPU.
With this configuration a forward pass takes on average 24 ms per image ($320\times320$ pixels) on a NVIDIA RTX 3090 and an AMD Ryzen7 3700X.

\section{Experiments}

\textbf{Dataset:} Building extraction and polygonization networks require ground truth polygonal annotations in order to be trained.
Therefore, we perform all our experiments using the CrowdAI Mapping Challenge dataset \cite{Mohanty:2018}, which is composed of over 280k satellite images of size $300 \times 300$ pixels for training and 60k images for testing. 
In order to avoid pooling issues in the backbone, we upsample the images to $320 \times 320$ pixels.
The dataset provides the polygon annotations in MS COCO format \cite{lin2014microsoft}. \smallskip

\textbf{Evaluation metrics:} We evaluate and compare the results of PolyWorld computing classical segmentation and detection metrics, such as Intersection over Union (IoU), and MS COCO \cite{lin2014microsoft} Average Precision (AP) and Average Recall (AR).
In order to evaluate the regularity of the extracted building contours, we also calculate the Max Tangent Angle Error \cite{girard2021polygonal}.
This metric compares the tangent angles of the predicted and ground truth polygons, penalizing building contours not aligned with the ground truth. 

In general, simple polygonization methods applied to the raster output of classical segmentation networks produce unregular polygons with a high amount of redundant vertices.
On the other hand, building extraction and polygonization methods tend to reduce the segmentation scores in favour of more regular and realistic footprints.
Since the goal of the proposed method is to generate high quality building polygons ready to be used on geographical applications, we introduce the \textit{complexity aware IoU (C-IoU)} metric computed as follows:

\begin{equation}
    \text{C-IoU}(A, \bar{A}) = \text{IoU}(A, \bar{A}) \cdot \left( 1 - \text{RD}(N_A, N_{\bar{A}}) \right)
\end{equation}

\noindent
where the first term $\text{IoU}(A, \bar{A})$ indicates the intersection over union between the predicted polygon raster mask $A$ and the ground truth segmentation $\bar{A}$.
The second term $\text{RD}(N_A, N_{\bar{A}}) = |N_A - N_{\bar{A}}| / (N_A + N_{\bar{A}})$ is the relative difference between the number of extracted vertices $N_A$ in the image used to produce the raster $A$, and the number of ground truth vertices $N_{\bar{A}}$.
The metric aims to favor polygonizations with a complexity similar to ground truth, penalizing both oversimplified building shapes and polygons with redundant vertices.
Ideally, a method achieves a high C-IoU score if it manages to balance the trade off between segmentation accuracy and polygonization complexity.

\begin{table*}
\centering
\resizebox{\linewidth}{!}{%
\begin{tabular}{l|cccccc|cccccc}
\hline
\textbf{Method}              & $AP$            & $AP_{50}$          & $AP_{75}$          & $AP_{S}$           & $AP_{M}$           & $AP_{L}$           & $AR$            & $AR_{50}$          & $AR_{75}$          & $AR_{S}$           & $AR_{M}$           & $AR_{L}$           \\ \hline
Mask R-CNN\cite{he2017mask}                   & 41.9          & 67.5          & 48.8          & 12.4          & 58.1          & 51.9          & 47.6          & 70.8          & 55.5          & 18.1          & 65.2          & 63.3          \\
PANet\cite{liu2018path}                        & 50.7          & 73.9          & 62.6          & 19.8          & 68.5          & 65.8          & 54.4          & 74.5          & 65.2          & 21.8          & 73.5          & 75.0          \\
PolyMapper\cite{li2019topological}                   & 55.7          & 86.0          & 65.1          & 30.7          & 68.5          & 58.4          & 62.1          & 88.6          & 71.4          & 39.4          & 75.6          & 75.4          \\ \hline
FFL (no field), mask          & 57.8          & 84.0          & 66.9          & 33.8          & 74.1          & 80.7          & 67.0          & 90.4          & 76.9          & 46.2          & 79.7          & 85.7          \\
FFL (no field), simple poly   & 61.1          & 87.4          & 71.2          & 35.1          & 74.5          & 82.3          & 64.7          & 89.4          & 74.1          & 41.7          & 77.9          & 85.7          \\
FFL (with field), mask        & 57.7          & 83.8          & 66.3          & 33.8          & 73.8          & 81.0          & 68.1          & 91.0          & 77.7          & 47.5          & 80.0          & 86.7          \\
FFL (with field), simple poly & 61.7          & 87.6          & \textbf{71.4} & 35.7          & 74.9          & 83.0          & 65.4          & 89.8          & 74.6          & 42.5          & 78.6          & 85.8          \\
FFL (with field), ACM poly\cite{girard2021polygonal}    & 61.3          & 87.4          & 70.6          & 33.9          & 75.1          & 83.1          & 64.9          & 89.4          & 73.9          & 41.2          & 78.7          & 85.9          \\ \hline
PolyWorld (offset off)       & 58.7          & 86.9          & 64.5          & 31.8          & 80.1          & 85.9          & 71.7          & 92.6          & 79.9          & 47.4          & 85.7          & 94.0          \\
PolyWorld (offset on)        & \textbf{63.3} & \textbf{88.6} & 70.5          & \textbf{37.2} & \textbf{83.6} & \textbf{87.7} & \textbf{75.4} & \textbf{93.5} & \textbf{83.1} & \textbf{52.5} & \textbf{88.7} & \textbf{95.2} \\ \hline
\end{tabular}
}
  \caption{MS COCO \cite{lin2014microsoft} results on the CrowdAI test dataset\cite{Mohanty:2018} for all the building extraction and polygonization experiments. The results of PolyWorld are calculated discarding the correction offsets (offset off), and refining the vertex positions (offset on). FFL refers to the Frame Field Learning \cite{girard2021polygonal} method. The results are computed with and without frame field estimation. ``mask" refers to the pure segmentation produced by the model. ``simple poly" refers to the Douglas–Peucker polygon simplification \cite{douglas1973algorithms}, and ``ACM poly" refers to the Active Contour Model \cite{girard2021polygonal} polygonization method.}
  \label{tab:coco}
\end{table*}

\begin{table}
\centering
\resizebox{\linewidth}{!}{%
\begin{tabular}{l|cccc}
\hline
\textbf{Method}               & IoU           & C-IoU         & MTA            & N ratio \\ \hline
FFL (no field), simple poly   & 83.9          & 23.6          & 51.8°          & 5.96    \\
FFL (with field), simple poly & 84.0          & 30.1          & 48.2°          & 2.31    \\
FFL (with field), ACM poly    & 84.1          & 73.7          & 33.5°          & 1.13    \\ \hline
PolyWorld (offset off)        & 89.9          & 86.9          & 35.0°          & 0.93    \\
PolyWorld (offset on)         & \textbf{91.3} & \textbf{88.2} & \textbf{32.9°} & 0.93    \\ \hline
\end{tabular}
}
  \caption{\textit{Intersection over union (IoU)}, \textit{max tangent angle error (MTA)}, and \textit{complexity aware IoU (C-IoU)} results on the test-set of the CrowdAI dataset\cite{Mohanty:2018}. The last column reports the ratio between the number of detected vertices and the number of ground truth vertices.}
  \label{tab:iou_MTA}

\end{table}


\textbf{Results:} Qualitative results of experiments conducted on the CrowdAI \cite{Mohanty:2018} dataset can be observed in Figure \ref{fig:thumbnails}.
The images represent different kind of urban areas and they are sorted by building complexity from left to right.
We compare the results of PolyWorld with the Frame Field Learning (FFL) method\cite{girard2021polygonal} that represents the state of the art on building extraction and polygonization.
Both FFL and PolyWorld generalize well in every kind of building, but PolyWorld produces overall cleaner and more linear geometries without developing undesired artifacts.
It is interesting to note that PolyWorld can better deal with hard object occlusions, estimating the position of the hidden corners and connecting them producing more regular and realistic footprints, as shown on the left image.
The robustness of the vertex detection and matching process is shown on the right images, where PolyWorld does not have issues in generating polygons of complex buildings with curved walls or inner courtyards.
More images can be found in the supplementary material.

In Table \ref{tab:coco} we report the MS COCO metrics results using the test-set of CrowdAI.
We computed the scores of PolyWorld considering and discarding the positional offsets used to correct the vertex positions (``offset on" and ``offset off").
Our approach is compared with FFL, PolyMapper \cite{li2019topological}, and two general-purpose instance segmentation networks: Mask R-CNN \cite{he2017mask} and PANet\cite{liu2018path}.
For the FFL method, we report the results of the model trained with and without frame field output, and with different polygonization approaches: ``mask" is raster segmentation, ``simple poly" refers to the marching squares \cite{lorensen1987marching} contour detector followed by the Douglas–Peucker \cite{douglas1973algorithms} simplification, and ``ACM poly" refers to the Active Contour Model \cite{girard2021polygonal} polygonization.
The results of PolyWorld show state-of-the-art precision and recall performances despite the fact that the refinement offsets have been ignored.
When the vertex position refinement is enabled, all the scores improve by a considerable margin, demonstrating the effectiveness of the refinement losses.
Another interesting fact to mention is that PolyWorld uses considerably fewer points to describe the buildings compared to the FFL approach.
In the 60k test images of CrowdAI dataset, the ground truth counts a total of about 4.4M vertices.
PolyWorld extracts 4.2M polygon vertices in the test-set, compared to the 5.1M extracted by FFL with ACM polygonization. 
Nevertheless, our approach is able to achieve better segmentation scores, suggesting that the PolyWorld vertex extraction is more efficient.

In Table \ref{tab:iou_MTA} we report the intersection over union, max tangent angle error, and complexity aware IoU results.
Again, there is a noticeable improvement in all the metrics exploiting the vertex position refinement.
Even though the ACM polygonization of FFL significantly outperforms the Douglas–Peucker polygonization in terms of MTA and C-IoU, the full PolyWorld method manages to overtake all the FFL results.

\section{Limitations and future work}
In our future work we want to demonstrate the capability of PolyWorld to generalize and produce accurate polygons on large scale data sets with a number of unseen conditions.
This will include the Inria segmentation dataset \cite{maggiori2017dataset} with Open Street Map annotations since it contains varied areas captured from different cities around the globe, and includes adjacent buildings with common corners.
From a technical point of view, the case of common corners could be efficiently solved using PolyWorld by generalizing the vertex detection network to multiclass segmentation, detecting the number of vertices located in the same position, and sampling the visual descriptor multiple times from the feature map if a shared corner is detected.
Another limitation of PolyWorld concerns buildings with holes.
Since the permutation matrix does not carry the information to bind outer and inner rings to the same shape, a post processing step might be required to generate multi-polygons.

\section{Conclusion}

We presented PolyWorld, a novel method capable of elegantly extracting building polygons from satellite and aerial images in an end-to-end manner.
The evaluation results experimentally prove the power and effectiveness of self-attention graph neural networks for matching and positional refinement of detected building vertices.
By solving an optimal transport problem, our method provides strong and reliable vertex connections and implicitly avoids redundant points.
Our experiments show that PolyWorld significantly outperforms existing building extraction approaches, enabling highly accurate and regular building footprints, which fulfill the strict requirements of geographic and cartographic applications.

\subsubsection*{Acknowledgments}
\footnotesize{Thanks to VRVis for financing the project.
VRVis is funded by BMK, BMDW, Styria, SFG, Tyrol and Vienna Business Agency in the scope of COMET - Competence Centers for Excellent Technologies (879730) which is managed by FFG.}

{\small
\bibliographystyle{ieee_fullname}
\bibliography{main}
}


\twocolumn[
\centering
\Large
\textbf{PolyWorld: Polygonal Building Extraction with Graph Neural Networks in Satellite Images} \\
\vspace{0.5em}Supplementary Material \\
\vspace{1.0em}
] 
\appendix

\normalsize

In the following pages, we present additional qualitative examples of PolyWorld applied to the CrowdAI test dataset \cite{Mohanty:2018}. 
In particular, we show a larger comparison with the state-of-the-art Frame Field Learning (FFL) approach \cite{girard2021polygonal}, additional results on challenging scenarios and on randomly sampled images from the CrowdAI test set, as well as failure cases of our approach.
Moreover, we show an ablation study to evaluate the individual components of our method.

\section*{Qualitative results}
\begin{figure}[t]
  \centering
    \includegraphics[width=0.99\linewidth]{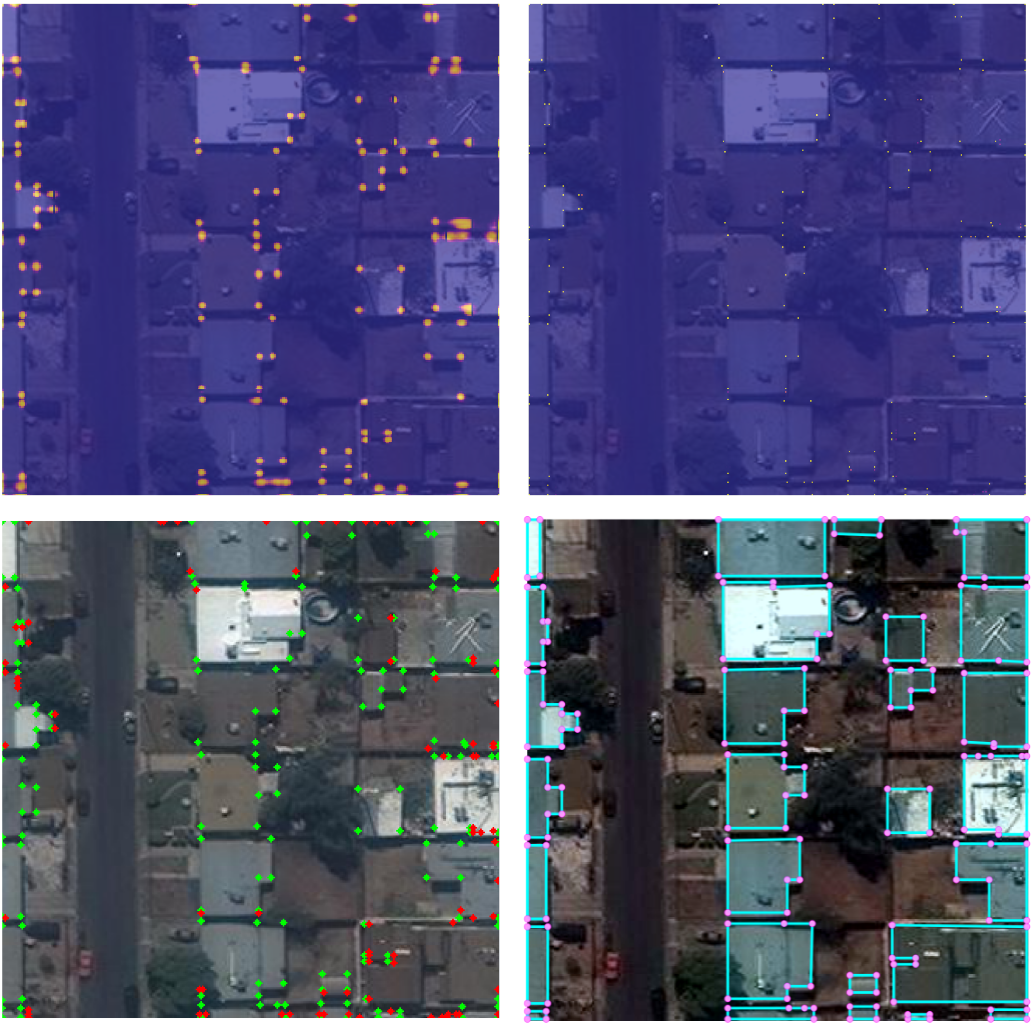}
  \caption{\textbf{Vertex detection heatmap.} Top-left: Probability map generated by the Vertex Detection Network. Top-right: probability map after Non Maximum Suppression (please zoom in for better view). Bottom-left: top-256 highest peaks. Green points indicate valid vertices, while red points indicate discarded vertices. Bottom-right: final result.}
  \label{fig:heatmap}
\end{figure}

\begin{figure*}[t]
  \centering
    \includegraphics[width=0.99\linewidth]{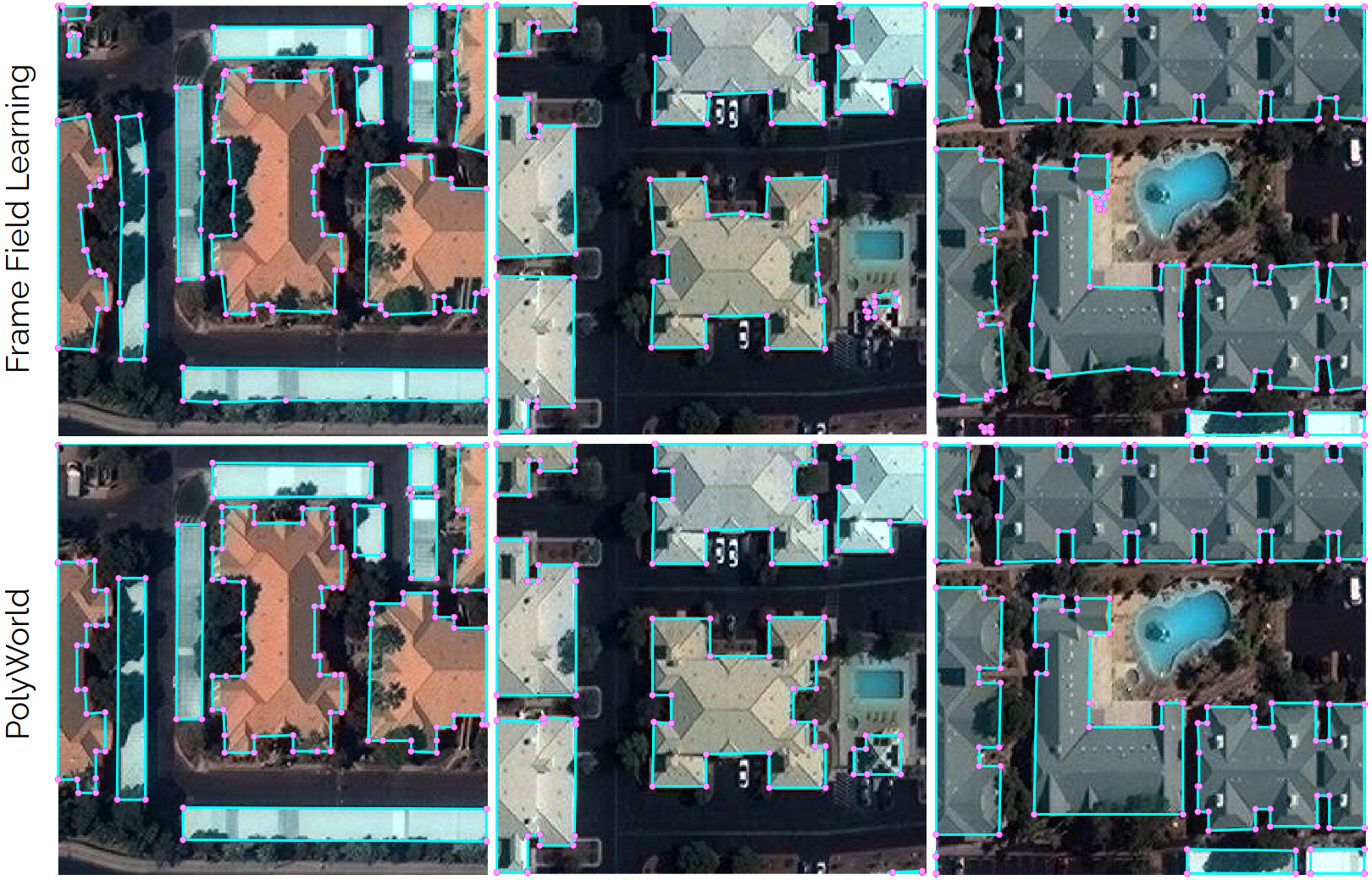}
  \caption{Examples of building extraction and polygonization on CrowdAI test dataset. Top row: Frame Field Learning (FFL) approach \cite{girard2021polygonal} with Res101-UNet as backbone and ACM polygonization. Bottom row: PolyWorld results.}
  \label{fig:pw_vs_ff}
\end{figure*}

\begin{figure*}[t]
  \centering
    \includegraphics[width=0.99\linewidth]{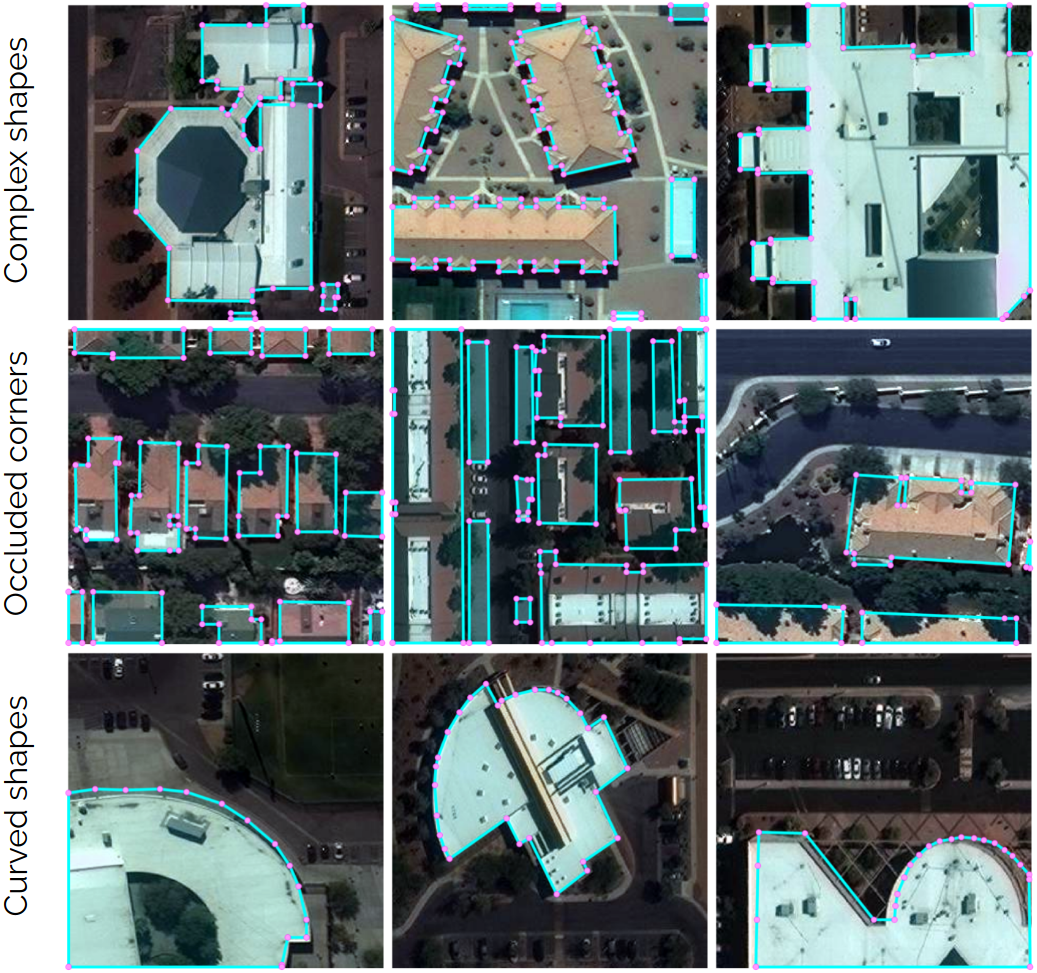}
  \caption{Results of PolyWorld on \textbf{challenging images} from the CrowdAI test dataset. First row: unusual and complex buildings. Second row: buildings with corners occluded by vegetation. Third row: buildings with curved walls.}
  \label{fig:hard}
\end{figure*}

\begin{figure*}[t]
  \centering
    \includegraphics[width=0.99\linewidth]{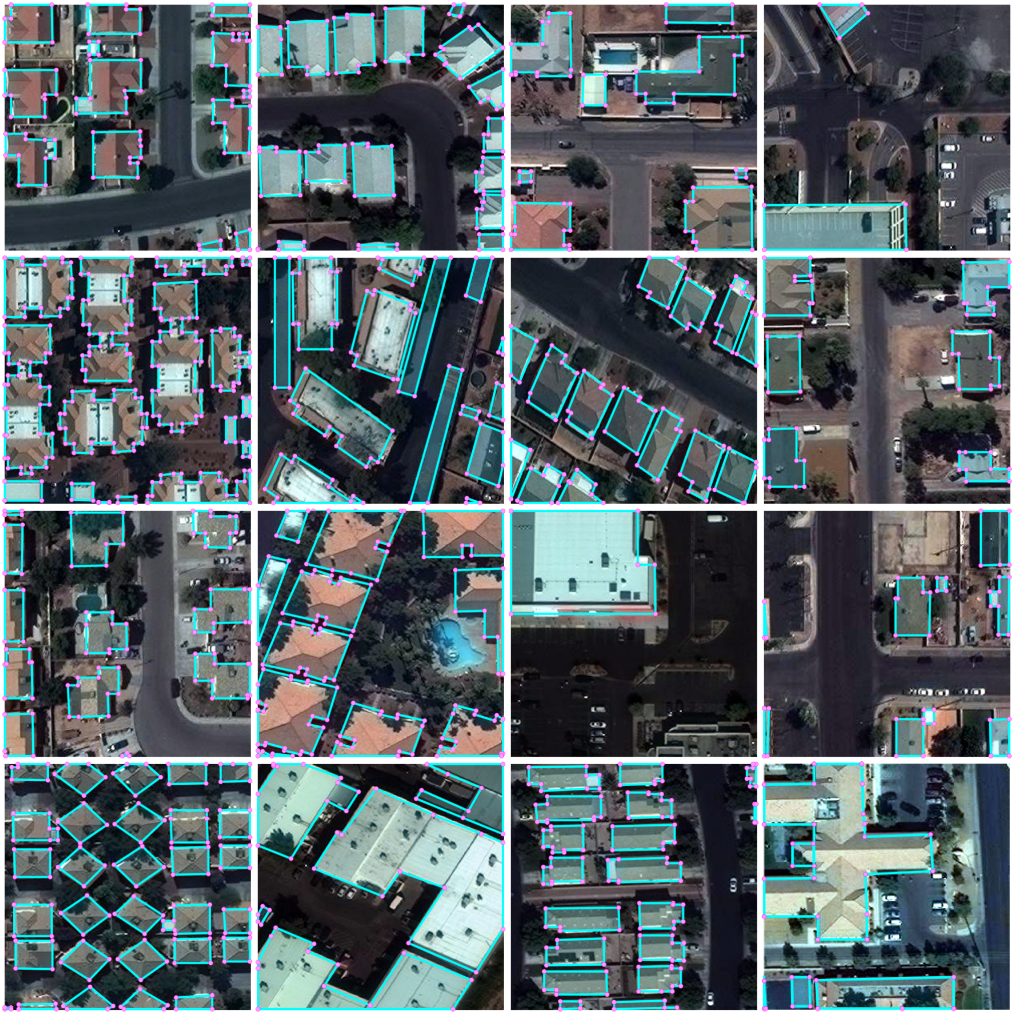}
  \caption{Results of PolyWorld on \textbf{randomly sampled images} from the CrowdAI test dataset.}
  \label{fig:random}
\end{figure*}

A qualitative comparison between PolyWorld and the Frame Field Learning (FFL) method are shown in Figure \ref{fig:pw_vs_ff}.
The images represent the results of the two different polygon extraction approaches on complex scenes selected from the CrowdAI test set.
Overall, PolyWorld utilizes a lower amount of vertices compared to FFL, generating more regular contours.
Results of our approach on challenging scenarios are shown in Figure \ref{fig:hard}.
PolyWorld demonstrates to generalize well on complex and unusual building shapes, managing to detect and connect precisely all the building corners also in presence of severe occlusions.
The vertex connections and the final polygon quality is noticeable even on buildings having curved walls as illustrated in the images in the bottom row of Figure \ref{fig:hard}. 
In Figure \ref{fig:random} additional PolyWorld polygonizations on randomly sampled images of the test set are shown. 
It is worth noting that some of the polygon predictions do not seem to be aligned with the building boundaries, especially on tall buildings.
This is caused by the fact that many images of the CrowdAI dataset are off nadir, but the annotations are aligned to the base of the buildings. 
The vertex detection and selection procedure is shown on a sampled CrowdAI test image in Figure \ref{fig:heatmap}.

\section*{Failure cases}
\begin{figure}[t]
  \centering
    \includegraphics[width=0.8\linewidth]{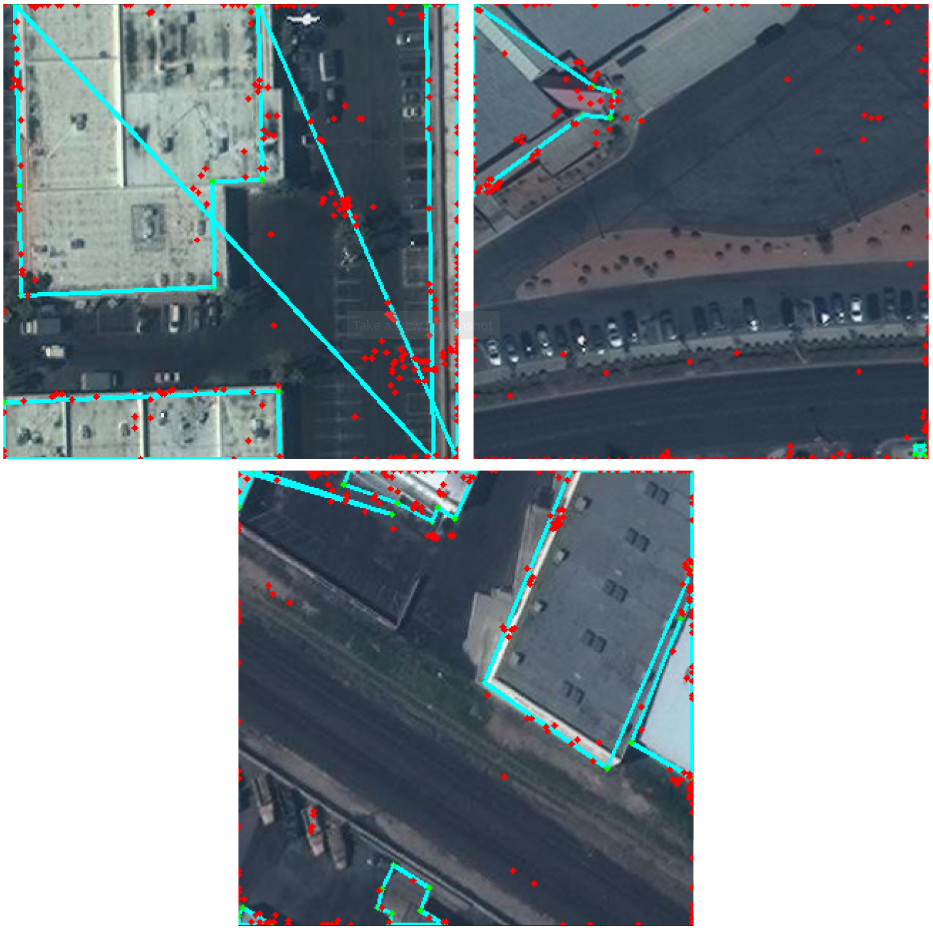}
  \caption{\textbf{Examples of wrong connections.} Green points indicate valid vertices. Red points indicate discarded vertices. Generated connections are shown in cyan.}
  \label{fig:failure}
\end{figure}

Even though PolyWorld experimentally proves to generate a reliable set of vertices and strong connections, it is interesting to show some failure cases caused by the optimal connection network.
In Figure \ref{fig:failure} we visualize three examples of wrong vertex matches, resulting from the linear sum assignment problem.
Red points describe vertices assigned to the diagonal of the permutation matrix and therefore are filtered, while valid vertices and connections are coloured in green and cyan, respectively.
On the left image two corners of the top-left building are assigned to the right building, generating an evident artifact.
In the right image the network discards some false-negative corners and does not complete the building footprint. 
Another artifact is generated by wrongly connecting two building corners to a false-positive vertex as shown in the bottom image.
These artifacts are very rare and therefore their impact in the segmentation performance is limited.

\begin{table*}
\centering
\resizebox{\linewidth}{!}{%
\begin{tabular}{l|c|c|cccccc|cccccc}
\hline
\textbf{PolyWorld} & \textbf{offset} & \textbf{score matrix $S$} & $AP$   & $AP_{50}$ & $AP_{75}$ & $AP_{S}$ & $AP_{M}$ & $AP_{L}$ & $AR$   & $AR_{50}$ & $AR_{75}$ & $AR_{S}$ & $AR_{M}$ & $AR_{L}$ \\ \hline
full method        & off             & $S_{clock} + S_{count}^\top$             & 58.7 & 86.9 & 64.5 & 31.8  & 80.1  & 85.9  & 71.7 & 92.6 & 79.9 & 47.4  & 85.7  & 94.0    \\
full method        & on              & $S_{clock} + S_{count}^\top$             & 63.3 & 88.6 & 70.5 & 37.2  & 83.6  & 87.7  & 75.4 & 93.5 & 83.1 & 52.5  & 88.7  & 95.2  \\\hline
full method        & on              & $S_{clock}$                     & 62.1 & 87.2 & 69.2 & 36.0    & 83.2  & 78.9  & 75.3 & 93.5 & 83.0   & 52.6  & 88.6  & 92.4  \\
full method        & on              & $S_{count}^\top$                   & 60.3 & 84.5 & 66.8 & 36.4  & 80.3  & 56.6  & 72.5 & 89.9 & 79.9 & 50.4  & 85.6  & 86.3  \\\hline
no GNN             & off (n/a)       & $S_{clock} + S_{count}^\top$             & 56.8 & 85.5 & 62.9 & 30.7  & 78.0    & 80.1  & 70.2 & 92.0   & 78.6 & 46.2  & 84.1  & 92.3  \\
no $\mathcal{L}_{angle}$      & on              & $S_{clock} + S_{count}^\top$             & 63.6 & 88.5 & 70.6 & 37.7  & 83.9  & 88.1  & 75.9 & 93.7 & 83.6 & 53.2  & 89.1  & 95.6  \\ \hline
\end{tabular}
}
  \caption{Ablation study. MS COCO \cite{lin2014microsoft} results on the CrowdAI test computed for different configurations of PolyWorld.}
  \label{tab:ablation_coco}
\end{table*}

\begin{table}
\centering
\resizebox{\linewidth}{!}{%
\begin{tabular}{c|c|c|ccc}
\hline
\textbf{PolyWorld} & \textbf{offset} & \textbf{score matrix $S$} & IoU & C-IoU & MTA  \\ \hline
full method        & off             & $S_{clock} + S_{count}^\top$             & 89.9         & 86.9           & 35.0°                   \\
full method        & on              & $S_{clock} + S_{count}^\top$             & 91.3         & 88.2           & 32.9°                    \\ \hline
full method        & on              & $S_{clock}$                     & 90.9         & 88.1           & 33.0°                    \\
full method        & on              & $S_{count}^\top$                   & 88.4         & 84.7           & 33.0°                    \\ \hline
no GNN             & off (n/a)       & $S_{clock} + S_{count}^\top$             & 89.2         & 86.3           & 35.3°                    \\
no $\mathcal{L}_{angle}$      & on              & $S_{clock} + S_{count}^\top$             & 91.4         & 88.6           & 34.0°                    \\ \hline
\end{tabular}
}
  \caption{Ablation study. \textit{Intersection over union (IoU)}, \textit{max tangent angle error (MTA)}, and \textit{complexity aware IoU (C-IoU)} results on the test-set of the CrowdAI dataset\cite{Mohanty:2018} computed for different configurations of PolyWorld.}
  \label{tab:ablation_iou_MTA}

\end{table}

\section*{Ablation study}
We conduct additional experiments to evaluate the performance contribution provided by different components of PolyWorld.
In particular:

\begin{itemize}
    \setlength\itemsep{0.1em}
    \item The model is evaluated discarding the offsets that refine the position of the vertices.
    \item The model is evaluated only using $S_{clock}$ or $S_{count}^\top$ as score matrix $S$, rather than the ensemble of the two.
    \item The model is retrained without using the GNN. Removing the GNN automatically means discarding the vertex offsets and the global aggregation of descriptors. In this case, the visual descriptors $\textbf{d}$ are directly used for matching.
    \item The model is retrained discarding the angle loss $\mathcal{L}_{angle}$. Only the segmentation loss $\mathcal{L}_{seg}$ is kept to learn the offsets.
\end{itemize}

Quantitative results of the ablation study are reported in Table \ref{tab:ablation_coco} and Table \ref{tab:ablation_iou_MTA}.

\begin{figure*}[t]
\centering

\begin{subfigure}{\linewidth}
  \centering
    \includegraphics[width=0.99\linewidth]{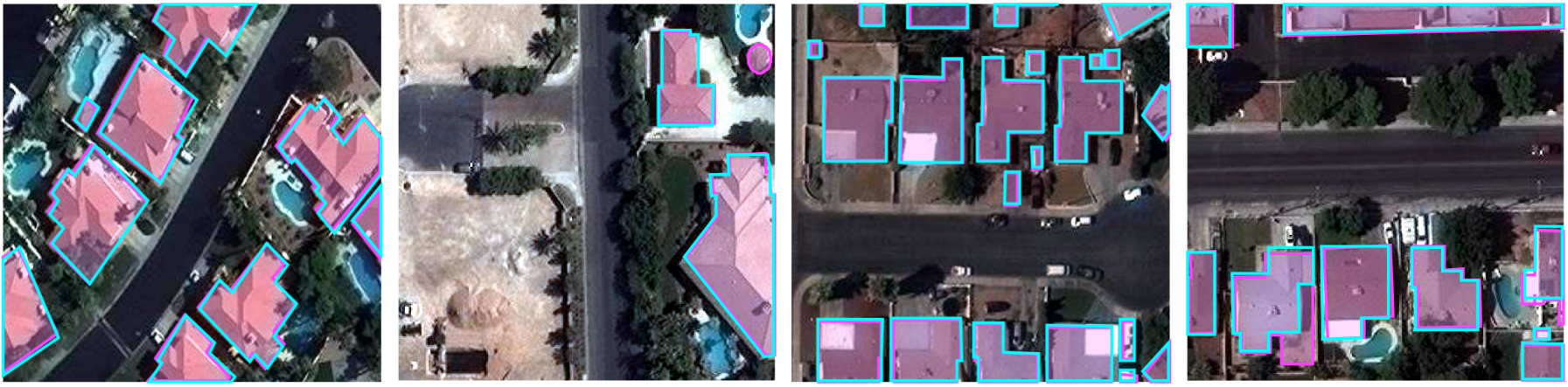}
  \caption{\textcolor{magenta}{\textbf{Magenta}}: Ground truth polygons. \textcolor{cyan}{\textbf{Cyan}}: Results of the full PolyWorld method.}
  \label{fig:gt_vs_full}
\end{subfigure}

\begin{subfigure}{\linewidth}
  \centering
    \includegraphics[width=0.99\linewidth]{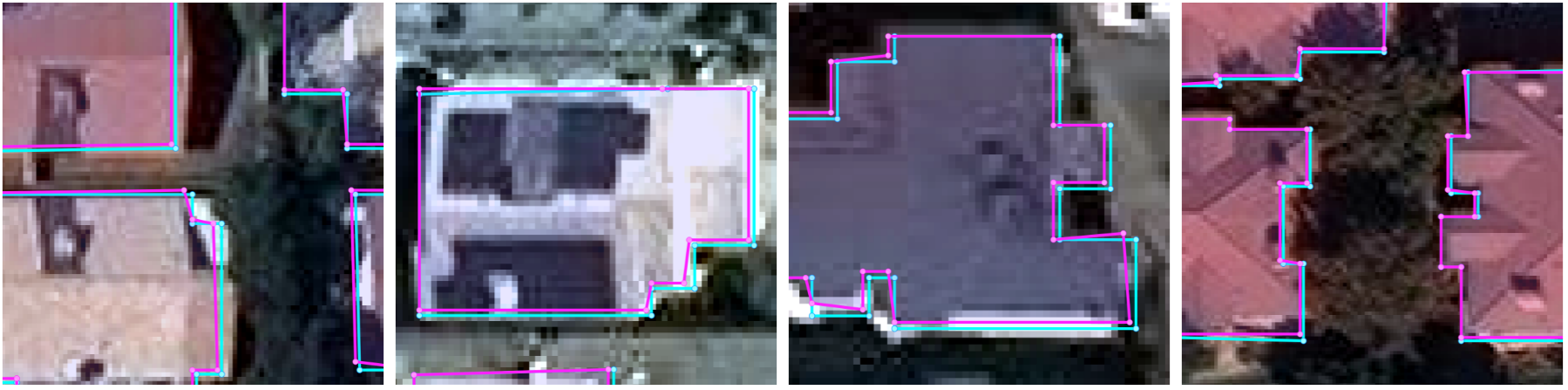}
  \caption{\textcolor{magenta}{\textbf{Magenta}}: PolyWorld results obtained discarding the refinement offsets. \textcolor{cyan}{\textbf{Cyan}}: Results of the full PolyWorld method.}
  \label{fig:offset_off_vs_full}
\end{subfigure}

\begin{subfigure}{\linewidth}
  \centering
    \includegraphics[width=0.99\linewidth]{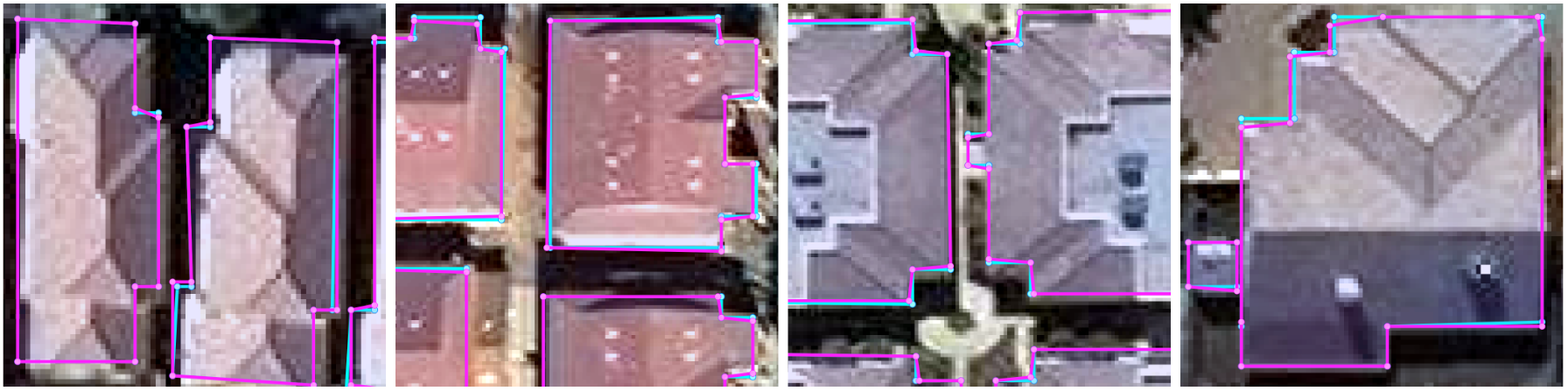}
  \caption{\textcolor{magenta}{\textbf{Magenta}}: PolyWorld results obtained discarding the angle loss $\mathcal{L}_{angle}$. \textcolor{cyan}{\textbf{Cyan}}: Results of the full PolyWorld method.}
  \label{fig:no_angle_loss_vs_full}
\end{subfigure}

\begin{subfigure}{\linewidth}
  \centering
    \includegraphics[width=0.99\linewidth]{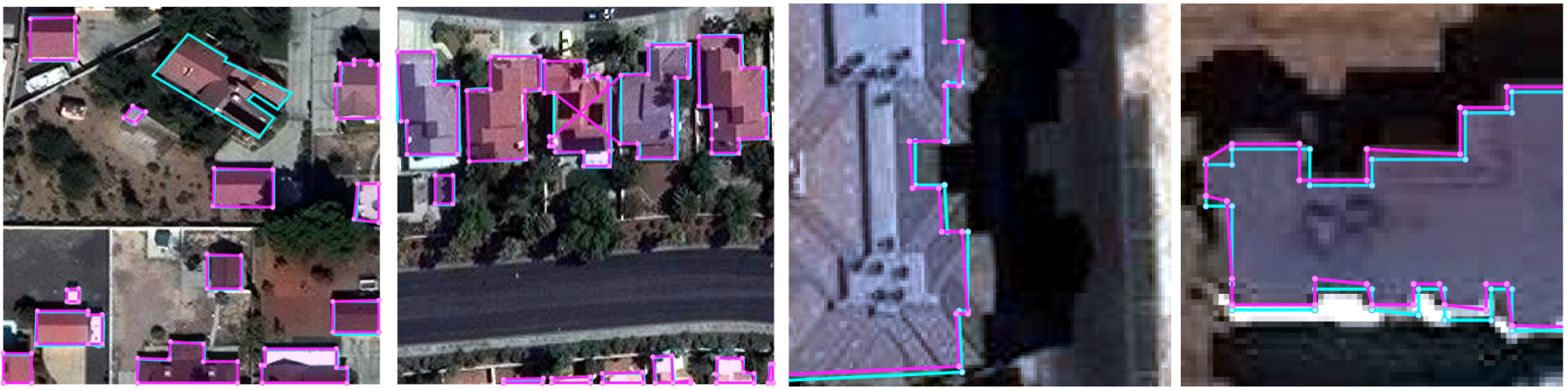}
  \caption{\textcolor{magenta}{\textbf{Magenta}}: PolyWorld results obtained discarding the Graph Neural Network. \textcolor{cyan}{\textbf{Cyan}}: Results of the full PolyWorld method.}
  \label{fig:noGNN_vs_full}
\end{subfigure}

\caption{Ablation study: qualitative results obtained with different configurations of PolyWorld.  }
\label{fig:ablation}
\end{figure*}

Discarding the refinement offsets results in a noticeable drop in detection and segmentation performance.
Moreover, the polygons visually appear not as regular as the full PolyWorld results, as shown in Figure \ref{fig:offset_off_vs_full}.

Equivalent or even higher detection scores than the full PolyWorld method are achieved retraining the model without angle loss.
Unfortunately, in this configuration PolyWorld is not encouraged to generate sharp building corners and the visual impact of the resulting building shapes is not as good as the polygons produced by the method trained with $\mathcal{L}_{angle}$, as shown in Figure \ref{fig:no_angle_loss_vs_full}.
This phenomenon is also explained by the fact that this configuration does not perform as well as the full method is terms of MTA (see Table \ref{tab:iou_MTA}).

Without GNN, PolyWorld still manages to make meaningful vertex connections even though it is not rare to encounter missing footprints or wrong matches, as shown in Figure \ref{fig:noGNN_vs_full}.

The quantitative results using $S = S_{clock}$ and $S = S^{\top}_{count}$ are reported in Table \ref{tab:coco} and Table \ref{tab:iou_MTA}.
As expected, only using $S_{clock}$ or $S^{\top}_{count}$ leads to lower segmentation scores compared to the combination of the two: $S = S_{clock} + S^{\top}_{count}$.

\section*{Runtimes}
The Frame Field Learning \cite{girard2021polygonal} paper reports a computation time of 0.04s on CrowdAI using a GTX 1080Ti. 
PolyWorld achieves a comparable computation time, taking 0.047s per image with the same configuration (or 0.024s on a RTX 3090).


\end{document}